\pdfoutput=1

\documentclass[11pt]{article}

\usepackage[]{ACL2023}

\usepackage{times}
\usepackage{latexsym}
\usepackage{inconsolata}
\usepackage{tabularx}
\usepackage{multirow}
\usepackage{longtable}
\usepackage{booktabs}
\usepackage{graphicx}
\usepackage[export]{adjustbox}
\usepackage{microtype}
\usepackage{algorithm}
\usepackage{algorithmic}
\usepackage{booktabs}
\usepackage{amsmath}
\usepackage{multirow}
\usepackage{bbding}
\usepackage{mathrsfs}
\usepackage{subfigure}
\usepackage{amssymb}
\usepackage{makecell}
\usepackage{graphicx}
\usepackage[T1]{fontenc}

\usepackage[utf8]{inputenc}

\usepackage{microtype}

\usepackage{inconsolata}

\title{DivTOD: Unleashing the Power of LLMs for Diversifying Task-Oriented Dialogue Representations}

\author{Weihao Zeng$^{1*}$, Dayuan Fu$^{1*}$, Keqing He$^{2}$, \\ { \bf Yejie Wang$^{1}$}, {\bf Yukai Xu$^{1}$}, {\bf Weiran Xu$^{1}$}\thanks{\ \ The first two authors contribute equally. Weiran Xu is the corresponding author.}\\
  $^1$Beijing University of Posts and Telecommunications, Beijing, China\\
$^{2}$Meituan, Beijing, China\\
  \texttt{\{zengwh,fdy,wangyejie,xuyukai,xuweiran\}@bupt.edu.cn}\\
  \texttt{\{hekeqing\}@meituan.com}
  }

\begin{document}
\maketitle
\begin{abstract}
Language models pre-trained on general text have achieved impressive results in diverse fields. Yet, the distinct linguistic characteristics of task-oriented dialogues (TOD) compared to general text limit the practical utility of existing language models. Current task-oriented dialogue pre-training methods overlook the one-to-many property of conversations, where multiple responses can be appropriate given the same conversation context.
In this paper, we propose a novel dialogue pre-training model called DivTOD, which collaborates with LLMs to learn diverse task-oriented dialogue representations. DivTOD guides LLMs in transferring diverse knowledge to smaller models while removing domain knowledge that contradicts task-oriented dialogues. Experiments show that our model outperforms strong TOD baselines on various downstream dialogue tasks and learns the intrinsic diversity of task-oriented dialogues. 
\end{abstract}

\section{Introduction}






Many NLP applications frequently utilize pre-trained language models (PLMs) \cite{devlin-etal-2019-bert, Liu2019RoBERTaAR}, which are based on extensive general text corpora \cite{Zhu2015AligningBA}.These models are pre-trained in a self-supervised manner and then fine-tuned for supervised downstream tasks. The Pretrain and Finetune paradigm has significantly improved the performance of various downstream tasks. Despite their success, most current research efforts focus on general documents such as Wikipedia, which have a large linguistic gap with dialogues, particularly task-oriented dialogues. Directly using these PLMs is not ideal and yields poor performance \cite{Rashkin2019TowardsEO}.

\begin{figure}[t]
 \centering
 \setlength{
 \abovecaptionskip}{-0.1cm}
\resizebox{0.48\textwidth}{!}{
 \includegraphics[scale=0.7]{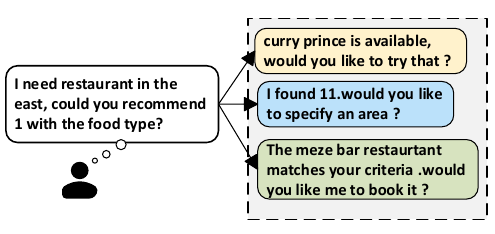}
 }

 \caption{The same context may have multiple appropriate responses in a task-oriented dialogue, which we call one-to-many.}

 \label{fig:intro}

\end{figure}

Compared to plain text, TOD aims to help users accomplish specific tasks with explicit goals (e.g. restaurant reservation), belief states, and database information. Thus, learning high-quality dialogue representations is crucial for understanding tasks in TOD. Previous methods pre-trained models using task-oriented dialogue datasets to improve dialogue understanding performance. SimCSE \cite{gao-etal-2021-simcse} uses a contrastive learning framework to learn sentence embeddings by generating positive pairs through Dropout \cite{Srivastava2014DropoutAS} augmentation. TOD-BERT \cite{Wu2020TODBERTPN} considers the intrinsic properties of dialogue data by using dialogue history and corresponding responses as positive pairs for contrastive learning. DSE \cite{Zhou2022LearningDR} learns from dialogues by taking consecutive utterances of the same dialogue as positive pairs. Furthermore, FutureTOD \cite{zeng2023futuretod} proposes a new non-contrastive self-training framework to address the challenges faced by previous contrastive methods in selecting true positive and negative pairs.

Despite previous TOD PLMs have made remarkable progress. Most work ignores the one-to-many property in the conversation where multiple responses can be appropriate under the same conversation context (shown in Figure \ref{fig:intro}). 
Our analysis shows that the lack of diversity in TOD datasets is the main reason for this. Specifically, (1) most TOD datasets only provide a single response for the same dialogue history, and (2) the style of system responses in TOD is often monotonous and dull. As a result, current TOD PLMs capture only the most common dialogue information and ignore less frequent but still feasible user behaviors, which leads to duplicated and plain responses.


Large Language Models (LLMs) \cite{Brown2020LanguageMA,Ouyang2022TrainingLM,Touvron2023LLaMAOA} offer hope for addressing the problems mentioned above. LLMs have more parameters and are pre-trained and fine-tuned on a richer and wider corpus \cite{kopf2023openassistant,chiangvicuna,ding2023enhancing}.  Consequently, LLMs possess a broader general background knowledge, which enables them to generate more diverse and feasible responses. However, it should be noted that LLMs have not been specifically fine-tuned for task-oriented dialogue systems \cite{hudevcek2023llms}, resulting in a significant mismatch between their general knowledge and the domain knowledge required for task-oriented dialogue. Furthermore, LLMs typically have billions of parameters, making them too expensive to deploy at scale because of the overwhelming computational requirements, as well as the cost of fine-tuning and inference \cite{wei2022chain}. To address these issues, a natural approach is to distill the rich background and domain-specific knowledge required for tasks from LLMs into smaller and more efficient models.

In this paper, we propose a new dialogue pre-training model, DivTOD, which enhances the ability of smaller models to model the intrinsic one-to-many diversity of human conversations by transferring rich general background knowledge and task-specific domain knowledge from LLMs. Our framework consists of three core steps: (1) Guiding LLMs to generate diverse system responses based on dialogue context in a "filling the blank" manner. (2) Using an LLM-based post-generation filter to align the generated responses with domain knowledge. (3) Allowing small models to imitate LLM's capabilities by observing diverse dialogues through self-training. We evaluated DivTOD on various task-oriented dialogue tasks, such as intent classification, dialogue state tracking, dialogue act prediction, and response selection. The results demonstrate that DivTOD consistently outperforms strong TOD baselines in all scenarios, indicating its generalization capability. 
Furthermore, we observed that DivTOD can capture a wider range of dialogue information and learn the intrinsic one-to-many diversity of TOD. 

Our contributions are:
(1) We propose a framework that distills task-specific domain knowledge and rich general background knowledge of LLMs into smaller models. We use this framework to pre-train DivTOD and model the intrinsic one-to-many diversity of human conversations.
(2) Our DivTOD outperforms strong TOD baselines on diverse downstream dialogue tasks. It also learns the intrinsic diversity of task-oriented dialogues

\section{Model}

 \begin{figure}[t]
 \centering
\resizebox{0.49\textwidth}{!}{
 \includegraphics[scale=0.5]{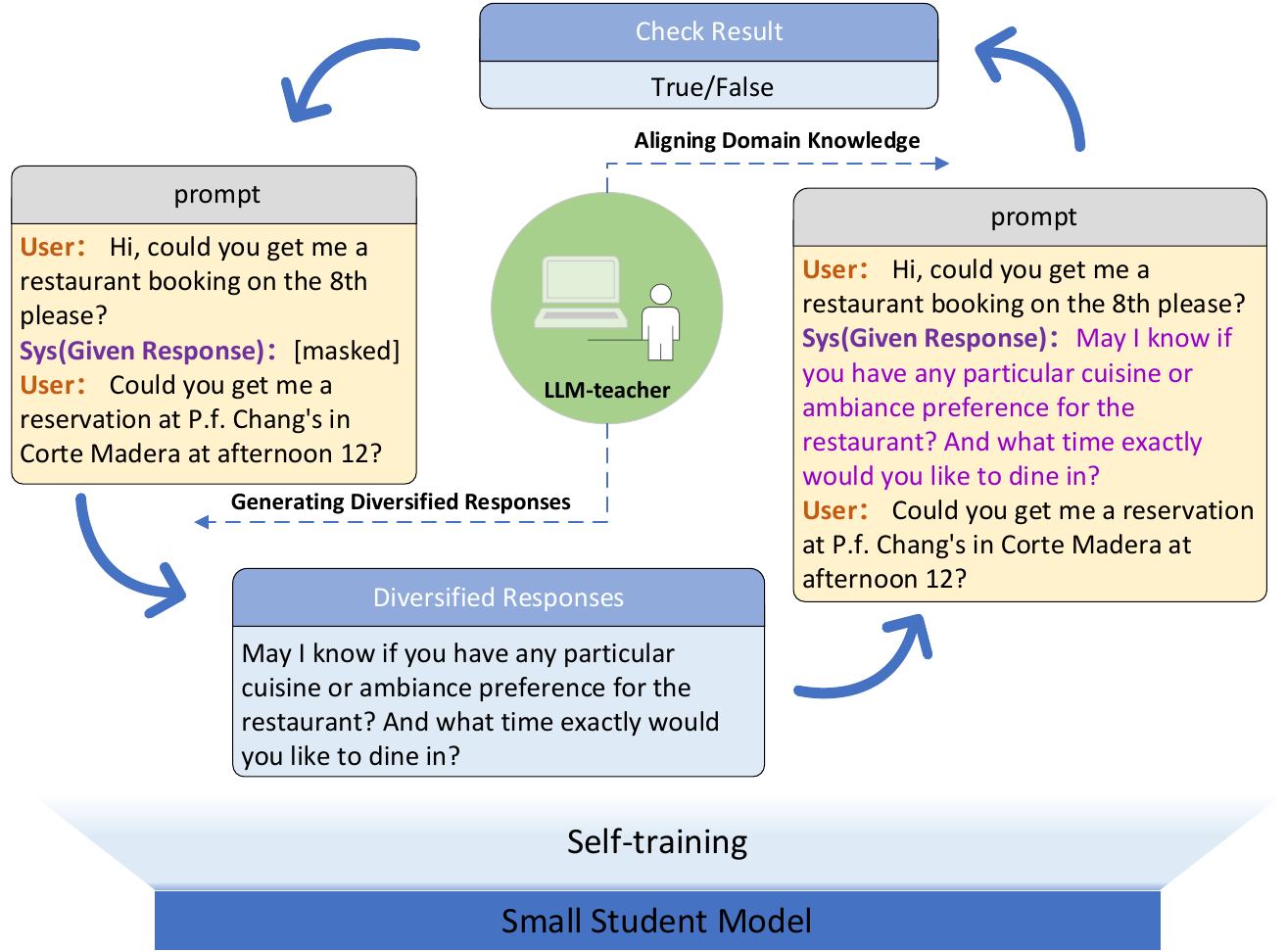}
 }

 \caption{Overall architecture of DivTOD.}
 \label{fig:model}

\end{figure} 
\subsection{Overall Architecture}

Figure \ref{fig:model} shows the overall architecture of DivTOD. Our framework comprises a teacher model $M_{T}$ based on LLM and a student model $M_{S}$ based on a smaller model, initialized by Vicuna-7b\footnote{https://huggingface.co/lmsys/vicuna-7b-delta-v1.1} and BERT-base-uncased\footnote{https://huggingface.co/bert-base-uncased}, respectively. First, we guide the $M_{T}$ to generate diverse system responses based on the dialogue context, using a "filling the blank" approach. Then, we use the $M_{T}$ as a filter to align the generated response with the domain knowledge of the task-oriented dialogue context. Finally, by continuously iterating the generate-filter steps, we enable the $M_{S}$ to train on both the original dataset and the generated dataset using the self-training method proposed in \citet{zeng2023futuretod}. 

\subsection{Diversifying Task-Oriented Dialogue
Representations}

\textbf{Notation}
We use the collected datasets by TOD-BERT \cite{Wu2020TODBERTPN} as our pre-training corpus. The corpus is the combination of 9 publicly available task-oriented datasets, including 100,707 dialogues and 1,388,152 utterances over 60 domains. For each dialogue, we first transform it into a token sequence $D=\left\{U_{1}, S_{1}, \ldots, U_{n}, S_{n}\right\}$. $U_{i}$ and $S_{i}$ denote the user utterance and system utterance with a prefix of two special role tokens [USR] or [SYS], respectively. $n$ is the turn number of the dialogue.

\textbf{Generating Diversified Responses} We use a "filling in the blank" approach to guide $M_{T}$ in generating diverse responses based on the dialogue context. For a given dialogue $D$, we randomly mask a system response $S_{i}$ and use the remaining part as the input $D^{'}$. We design a few-shot prompt $P$ consisting of a triplet $(I^{P}, D^{P}, S^{P})$ to instruct $M_{T}$ in generating diverse responses $S_{i}^{'}$ based on $D^{'}$ \footnote{We try different methods to instruct $M_{T}$, including zero-shot prompts. However, these methods are not very effective. For example, the pass rate of the zero-shot method is low in our post-filter.}. The core component of the $P$ is $I^{P}$, which describes the task to the model. $I^{P}$ also constrains the behavior of the model, preventing it from generating irrelevant responses. $D^{P}$ is the input example, and $S^{P}$ is the corresponding generated response \footnote{We have manually created a set of few-shot examples. During the generation process, we will randomly select samples to include in $D^{P}$ and $S^{P}$.}. For each input $D^{'}$, we append it to $P$ and use it as input to prompt $M_{T}$ to generate diverse responses. $M_{T}$ can mimic the demonstrations $D^{P},S^{P}$ in $P$ to generate new diverse responses. The complete prompt example is shown in Figure \ref{fig:prompt_gen} in Appendix. 


\textbf{Aligning Domain Knowledge} Although we can obtain more diverse responses using $M_{T}$, these responses may contradict the characteristics of task-oriented dialogue systems. For example, the generated responses may provide excessive information that the user does not need or answer questions that the user asks in the future. To ensure that the generated responses align with domain knowledge in TOD, we designed a filter based on $M_{T}$.We replace the masked parts in $D^{'}$ with the generated response $S_{i}^{'}$ to form a new input $D^{''}$. We have designed a few-shot prompt $E$ consisting of a triplet $(I^{E}, D^{E}, R^{E})$ to prompt $M_{T}$ to judge the contextual consistency of $D^{''}$ and whether it conflicts with the characteristics of TOD. The core part of the prompt is $I^{E}$, which describes the task to $M_{T}$. The prompt also provides logical knowledge related to task-oriented dialogue. $D^{E}$ and $R^{E}$ are the demonstrations provided to $M_{T}$. $D^{E}$ represents the example input, and $R^{E}$ represents the corresponding judgment result (either True or False). We append $D^{''}$ to $E$ and determine whether to keep $S_{i}^{'}$ based on the filtering result. Figure \ref{fig:prompt_eval} in the Appendix shows a complete example of this prompt.

\textbf{Self Training} We iterate through the generate-filter steps (summarized in Algorithm \ref{algo} in Appendix) described above and combine the newly generated dialogues with the original ones. 

We train $M_{S}$ using the self-training objective proposed by FutureTOD \cite{zeng2023futuretod} on the assembled dialogues. 
We initialize the new student model and teacher model using $M_{S}$ \footnote{The Student model and Teacher model here differ from the previously mentioned $M_{T}$ and $M_{S}$, which are concepts involved in self-training \cite{zeng2023futuretod}.}. For each dialogue, we randomly split it into context and future sequences. The student model encodes the context and obtains the original dialogue representation, while the teacher model encodes both the context and future to obtain the target. The architectures of the student and teacher models are the same, but the weights of the teacher model are periodically updated by the student. The training goal is to align the original content representation with the full representation containing future knowledge. The generate-filter steps produce diverse responses, resulting in multiple reasonable full representations that can align with the same content representation.

Through the above framework of generation, filtering, and self-training, we transfer both general background knowledge and task-specific domain knowledge from $M_{T}$ to $M_{S}$. 

\section{Experiment}

\subsection{Pre-training Corpus}

We use the nine different task-oriented datasets collected by \citet{Wu2020TODBERTPN} and show the full details in Appendix \ref{sec:data_stastics}.

\subsection{Baselines}
DivTOD is evaluated on various downstream tasks and compared to several well-established baselines, including both encoder-only and generative architectures. For details about the baselines, please refer to the Appendix \ref{sec:baseline}. 

\subsection{Implementation Details}
\textbf{LLM generating Details} 
We use Vicuna as LLM to generate diverse system responses and to align domain knowledge.
For experimental details and hyperparameter settings for this stage, please refer to the Appendix \ref{sec:llm_gen_detail}.

\textbf{Pre-training Details} After diverse system response generation, all the dialogue will be merged into the original dataset as the new dataset to pre-train. The details of the hyperparameters for the pre-training can be found in the Appendix \ref{sec:pre_train_detail}.

\textbf{Finetuning Details}  After completing pre-training on dialogue, we perform supervised fine-tuning on downstream dialogue tasks. However, it is important to note that we only use generated diverse dialogue during the pre-training phase. In the fine-tuning phase, we use datasets and settings that are identical to the previous baseline, including golden labels such as dialogue acts. 
  The details of the hyperparameters for the pre-training can be found in the Appendix \ref{sec:fine_tune_detail}.

\vspace{-0.2cm}
\subsection{Main Results}

We evaluate all the pre-trained LMs on four core task-oriented dialogue tasks: intent recognition, dialogue state tracking, dialogue act prediction, and response selection. It is important to emphasize that our focus is on learning diverse dialogue representations. Therefore, we are more concerned with tasks related to dialogue understanding rather than tasks related to response generation.
To ensure fairness in our evaluation, we adopt the same architecture for all baselines following TOD-BERT and only add simple components to the pre-trained model, such as a single-layer classification head. For each downstream task, we conduct experiments using the entire dataset. In addition, we also explored few-shot setting experiments in section \ref{sec:few_shot_setting}. This allowed us to see how well these pre-trained language models generalize to multiple tasks and scenarios.

\textbf{Intent Recognition} is a multi-class classification task that takes a dialogue utterance as input and predicts an intent label \cite{zeng2022semi}. We use the [CLS] embeddings from the model as the dialogue representation. We use cross-entropy loss to train the model. We report classification accuracy and recall.

The results of intent recognition on the OOS dataset \cite{larson-etal-2019-evaluation}, which encompasses 151 intent classes across ten domains, including 150 in-domain intents and out-of-domain (OOD) intents, are displayed in Table \ref{main_intent}. We find DivTOD outperforms all the baselines on 3 of 4 metrics, especially with significant improvements in overall accuracy and OOD metrics. All the results show the generalization ability of DivTOD both on in-domain and out-of-domain metrics.

\begin{table}[t]
\centering
\resizebox{0.50\textwidth}{!}{
\begin{tabular}{l|cccc}
\hline
\multicolumn{1}{c|}{\textbf{Model}} & \textbf{ACC(acc)} & \textbf{Acc(in)} & \textbf{Acc(out)} & \textbf{Recall(out)} \\ \hline
BERT                                & 84.9\%            & 95.8\%           & 88.1\%            & 35.6\%               \\
DialoGPT                            & 83.9\%            & 95.5\%           & 87.6\%            & 32.1\%               \\
BERT-mlm                            & 85.9\%            & 96.1\%           & 89.5\%            & 46.3\%               \\
SimCSE                              & 82.3\%            & 94.7\%           & 86.6\%            & 26.6\%               \\
TOD-BERT                            & 86.6\%            & \textbf{96.2\%}  & 89.9\%            & 43.6\%               \\
DSE                                 & 84.3\%            & 95.8\%           & 87.7\%            & 32.5\%               \\
FutureTOD                           & 87.2\%            & 96.0\%           & 90.0\%            & 47.6\%               \\
DivTOD                     & \textbf{87.4\%}*   & 95.8\%           & \textbf{90.5\%}*   & \textbf{49.5\%}*      \\ \hline
\end{tabular}
}

\caption{Intent recognition results on the OOS dataset. Acc(all), Acc(in), Acc(out) denotes the overall accuracy, in-domain intent accuracy, and out-of-domain intent accuracy. The numbers with * are significant using t-test with $p < 0.01$.}
\label{main_intent}

\end{table}

\textbf{Dialogue State Tracking} is also a multi-class classification task, which involves identifying the slot values for each (domain, slot) pair at each dialogue turn, based on a pre-defined ontology. The model takes dialogue history as input and is trained with cross-entropy loss summed over all the pairs. We employ the commonly used TOD dataset MWOZ 2.1 \cite{Budzianowski2018MultiWOZA}, which spans seven distinct domains, and we present the Joint acc and Slot acc. The Joint acc is deemed true solely if the predicted values align with their ground truth values at each dialogue turn. The Slot acc, on the other hand, independently contrasts each (domain, slot, value) triplet with its corresponding ground truth label.

Table \ref{main_dst} shows the results of dialogue state tracking on MWOZ 2.1. Our DivTOD excels by achieving the best results across all metrics. We find SimCSE performs poorly because it ignores the intrinsic properties of dialogue data and can not model overall dialogue. Our method achieves a greater improvement on joint accuracy than on slot accuracy, indicating the strength of understanding the overall dialogue context. For example, DivTOD outperforms TOD-BERT by 0.3\% on Slot Acc but 2.9\% on Joint Acc in the full data setting, which indicates the superiority of dialogue modeling.

\begin{table}[t]
\centering
\resizebox{0.3\textwidth}{!}{
\begin{tabular}{l|cc}
\hline
\multicolumn{1}{c|}{\textbf{Model}} & \textbf{Joint Acc} & \textbf{Slot Acc} \\ \hline
BERT                                & 45.6\%            & 96.6\%           \\
BERT-mlm                            & 47.7\%            & 96.8\%           \\
SimCSE                              & 48.0\%            & 96.8\%           \\
TOD-BERT                            & 48.0\%            & 96.9\%           \\
DSE                                 & 49.9\%            & 97.0\%           \\
FutureTOD                           & 50.4\%            & 97.1\%           \\
DivTOD                     & \textbf{50.9\%}*   & \textbf{97.2\%}*  \\ \hline
\end{tabular}}

\caption{Dialogue state tracking results on MWOZ 2.1. We report joint goal accuracy (Joint Acc) and slot accuracy (Slot Acc). The numbers with * are significant using t-test with $p < 0.01$.}

\label{main_dst}
\end{table}

\textbf{Dialogue Act Prediction}
is a multi-label classification task that uses dialogue history as input to forecast multiple dialogue acts related to the system response. The model employs binary cross-entropy loss across all potential actions for training. At the time of inference, the threshold to activate the dialogue act is established at 0.5. We use two datasets MWOZ \cite{Budzianowski2018MultiWOZA} and DSTC2 \cite{Henderson2014TheSD}. Following \cite{Wu2020TODBERTPN}, We implement identical data preprocessing to standardize the original dialogue acts into a universal format. We present the micro-F1 and macro-F1 results.

Table \ref{main_act} displays dialogue act prediction's result on MWOZ and DSTC2 datasets.
Our DivTOD method outperforms all other baselines in three out of four metrics. Specifically, our method surpasses FutureTOD on the DSTC2 dataset, demonstrating significant improvement. It also exhibits improvement on MWOZ, with the macro-F1 increasing from 81.9\% to 82.6\%. However, we notice that different methods exhibit unclear distinctions in terms of micro-F1. We attribute this to the imbalanced distribution of dialogue action labels in MWOZ. In such cases, macro-F1 provides a more reasonable evaluation metric as it assigns equal weight to each label, regardless of the number of samples.
In addition to the higher response quality, we also observe that DivTOD captures a wider range of dialogue policies and learns the intrinsic one-to-many diversity of TOD, as discussed in Section \ref{sec:resp_diversity}.
\begin{table}[t]
\centering
\resizebox{0.50\textwidth}{!}{
\begin{tabular}{l|cc|cc}
\hline
\multicolumn{1}{c|}{\multirow{2}{*}{\textbf{Model}}} & \multicolumn{2}{c|}{\textbf{MWOZ}} & \multicolumn{2}{c}{\textbf{DSTC2}}         \\ \cline{2-5} 
\multicolumn{1}{c|}{}                       & \textbf{micro-F1}    & \textbf{macro-F1}   & \textbf{micro-F1}        & \textbf{macro-F1}        \\ \hline
BERT                                        & 91.4\%      & 79.7\%      & 92.3\%          & 40.1\%          \\
DialoGPT                                    & 91.2\%      & 79.7\%      & 93.8\%          & 42.1\%          \\
BERT-mlm                                    & 91.7\%      & 79.9\%      & 90.9\%          & 39.9\%          \\
SimCSE                                      & 91.6\%      & 80.3\%      & 91.5\%          & 39.6\%          \\
TOD-BERT                                    & 91.7\%      & 80.6\%      & 93.8\%          & 41.3\%          \\
DSE                                         & 91.7\%      & 81.3\%      & 92.6\%          & 40.2\%          \\
FutureTOD                                   & \textbf{92.0\%}      & 81.9\%      & 94.6\%          & 44.6\%          \\
DivTOD                             & 91.7\%      & \textbf{82.6\%}*      & \textbf{95.8\%}* & \textbf{46.5\%}* \\ \hline
\end{tabular}
}

\caption{Dialogue act prediction results on MWOZ and DSTC2. The numbers with * are significant using t-test with $p < 0.01$.}
\label{main_act}

\end{table}

\textbf{Response Selection} is a ranking task that aims to retrieve the most relative system response from a candidate pool based on dialogue history. We also use MWOZ and DSTC2 as our evaluation datasets. We use a dual-encoder strategy, which calculates cosine similarity scores between dialogue history and candidate responses. We train this model with random system responses from the corpus as negative samples. We report k-to-100 accuracy. This metric represents the ratio of the ground-truth response being ranked in the top-k positions when compared to 99 randomly sampled responses, as determined by the scores computed by the dual-encoder.

Table \ref{main_rs} displays the results of response selection on MWOZ and DSTC2. Our DivTOD method achieves state-of-the-art results on all metrics. Despite TOD-BERT being pre-trained with a response contrastive objective, our method still significantly outperforms it on both MWOZ and DSTC2 in full data settings. This indicates that Our Method has better generalization capabilities. Compared to FutureTOD, our method brings significant improvements in response selection, indicating that it can enhance the diversity of TOD representation and thus improve performance. As the context representation of the pre-trained model becomes richer, it can retrieve more precise responses, as reflected in top-1, top-3, top-10 accuracy.

In summary, our method shows notable improvements in dialogue act prediction and response selection tasks. This indicates that considering the one-to-many nature of dialogues is essential for these tasks. Furthermore, our method also achieves enhancement in other important task-oriented dialogue tasks, such as intent classification and dialog state tracking. This further highlights the generalization of our method across various tasks.


\begin{table}[t]
\centering
\resizebox{0.46\textwidth}{!}{
\begin{tabular}{l|cc|cc}
\hline
\multicolumn{1}{c|}{\multirow{2}{*}{\textbf{Model}}} & \multicolumn{2}{c|}{\textbf{MWOZ}}    & \multicolumn{2}{c}{\textbf{DSTC2}}    \\ \cline{2-5} 
\multicolumn{1}{c|}{}                                & \textbf{1-to-100} & \textbf{3-to-100} & \textbf{1-to-100} & \textbf{3-to-100} \\ \hline
BERT                                                 & 47.5\%            & 75.5\%            & 46.6\%            & 62.1\%            \\
DialoGPT                                             & 35.7\%            & 64.1\%            & 39.8\%            & 57.1\%            \\
BERT-mlm                                             & 48.1\%            & 74.3\%            & 50.0\%            & 65.1\%            \\
SimCSE                                               & 64.2\%            & 85.4\%            & 55.6\%            & 70.5\%            \\
TOD-BERT                                             & 65.8\%            & 87.0\%            & 56.8\%            & 70.6\%            \\
DSE                                                  & 63.3\%            & 85.3\%            & 58.3\%            & 72.0\%            \\
FuturueTOD                                           & 68.5\%            & 87.9\%            & 58.4\%            & 72.6\%            \\
DivTOD                                      & \textbf{71.3\%}*   & \textbf{90.4\%}*   & \textbf{59.5\%}*   & \textbf{74.0\%}*   \\ \hline
\end{tabular}
}

\caption{Response selection evaluation results on MWOZ and DSTC. We report 1-to-100 and 3-to-100 accuracy, which represents the ratio of the ground-truth response being ranked at the top-1 or top-3 given 100 candidates. The numbers with * are significant using t-test with $p < 0.01$.}
\label{main_rs}

\end{table}

\section{Qualitative Analysis}
\subsection{Ablation Study of Domain Knowledge Alignment}
Table \ref{Filtering mechanism} presents the ablation study results of the domain knowledge alignment, on the two downstream tasks, dialogue act prediction on DSTC2 and response selection on MWOZ \footnote{Considering the cost of qualitative analysis, we select two classic task-oriented tasks. Furthermore, we use different datasets for each task to ensure generalizability.}. DivTOD performs the best under various conditions when training using dialogue with aligned domain knowledge. However, the performance of DivTOD w/o Align is unsatisfactory. For example, in the dialogue act prediction task, DivTOD w/o Align is similar to the baseline and lower than DivTOD's performance.
This suggests that aligning with domain knowledge may help maintain consistency in TOD dialogues, thereby contributing to the likelihood that the diverse dialogues generated by LLM have a beneficial influence on the pre-training process. 

\begin{figure*}[t]
 \centering
\resizebox{0.95\textwidth}{!}{
 \includegraphics[scale=0.5]{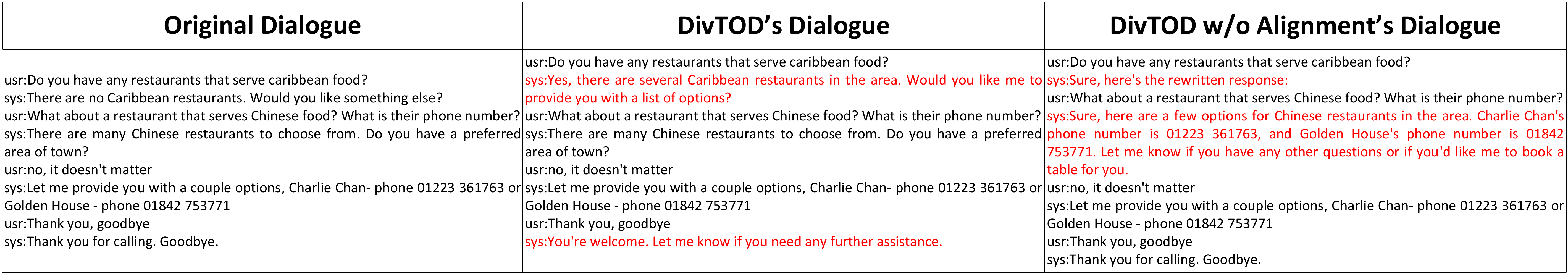}
 }

 \caption{Different Dialogue Cases. Original Dialogues refers to the dialogues from the original TOD dataset. DivTOD's Dialogue refers to the dialogues generated using the complete generating and aligning steps. DivTOD w/o Alignment's Dialogue refers to the dialogues generated after removing domain knowledge alignment.}
 \label{fig:case}

\end{figure*}

To visually represent the quality of the generated dialogues by different methods, we randomly selected dialogue samples, as shown in the Figure \ref{fig:case}. From the dialogue examples, it can be seen that the DivTOD's dialogues are different from the original text and they are all consistent with the dialogue context. However, DivTOD w/o Alignment's Dialogue produces two problems. First, LLM may not answer according to the prompt instructions, but may produce irrelevant answers such as “here’s the rewritten response:”. Second, LLM may produce answers that do not match the context, such as answering questions that users will only raise or provide information in the future.

\vspace{-0.2cm}
\subsection{Advantages of LLMs in Generating Diversified Responses}
\vspace{-0.1cm}
To demonstrate the advantage of LLMs over other models trained solely on TOD data in generating diversified responses, we randomly sample 500 TOD dialogue samples and generate responses using both PPTOD \cite{su2021multi} and LLM \footnote{We do not include PPTOD in the main results because they use supervised labels. We focus on unsupervised dialogue pretraining and do not compare with it for fairness.}. We compare the number of unique n-grams contained in the generated responses. Tabel \ref{tab:uniq_gram} demonstrates that the responses generated by LLM contain more unique n-grams than those generated by PPTOD, even surpassing the number of unique n-grams present in the original dialogue. We analyze that PPTOD, being pre-trained on the TOD dataset, overfits the limitations of that dataset, resulting in a decrease in the diversity of responses it generates. This further supports the evidence that LLM is capable of generating more diverse responses.

\begin{table}[t]
\centering
\renewcommand{\arraystretch}{0.5}
\resizebox{0.48\textwidth}{!}{
\begin{tabular}{l|c|c|c}
\hline
\multicolumn{1}{c|}{\textbf{Source}} & \textbf{Unique 1-gram} & \textbf{Unique 2-gram} & \textbf{Unique 3-gram} \\ \hline
Raw Dialogue                         & 380                    & 1512                   & 2391                   \\
PPTOD                                & 221                    & 736                    & 1189                   \\
Vicuna-7b                            & 357                    & 1793                   & 3497                   \\ \hline
\end{tabular}
}

\caption{The number of unique n-grams contained in the generated responses. "Raw Dialogue," "PPTOD," and "Vicuna-7b" refer to responses from the original dialogues, PPTOD, and the LLM we utilized.}
\label{tab:uniq_gram}

\end{table}


\begin{table}[t]
\centering
\renewcommand{\arraystretch}{0.5}
\resizebox{0.47\textwidth}{!}{
\begin{tabular}{|l|cc|cc|}
\hline
\multicolumn{1}{|c|}{\multirow{4}{*}{\textbf{Method}}} & \multicolumn{2}{c|}{\multirow{2}{*}{\textbf{MWOZ}}} & \multicolumn{2}{c|}{\multirow{2}{*}{\textbf{DSTC2}}} \\
\multicolumn{1}{|c|}{} & \multicolumn{2}{c|}{} & \multicolumn{2}{c|}{} \\ \cline{2-5} 
\multicolumn{1}{|c|}{} & \multicolumn{1}{c|}{\multirow{2}{*}{\textbf{1-to-100}}} & \multirow{2}{*}{\textbf{3-to-100}} & \multicolumn{1}{c|}{\multirow{2}{*}{\textbf{micro-F1}}} & \multirow{2}{*}{\textbf{macro-F1}} \\
\multicolumn{1}{|c|}{} & \multicolumn{1}{c|}{} &  & \multicolumn{1}{c|}{} &  \\ \hline
\multirow{2}{*}{FutureTOD} & \multicolumn{1}{c|}{\multirow{2}{*}{68.5\%}} & \multirow{2}{*}{87.9\%} & \multicolumn{1}{c|}{\multirow{2}{*}{94.6\%}} & \multirow{2}{*}{44.6\%} \\
 & \multicolumn{1}{c|}{} &  & \multicolumn{1}{c|}{} &  \\
\multirow{2}{*}{DivTOD w/o Align} & \multicolumn{1}{c|}{\multirow{2}{*}{70.0\%}} & \multirow{2}{*}{90.1\%} & \multicolumn{1}{c|}{\multirow{2}{*}{94.8\%}} & \multirow{2}{*}{44.2\%} \\
 & \multicolumn{1}{c|}{} &  & \multicolumn{1}{c|}{} &  \\ 
\multirow{2}{*}{DivTOD} & \multicolumn{1}{c|}{\multirow{2}{*}{\textbf{71.3\%}}} & \multirow{2}{*}{\textbf{90.4\%}} & \multicolumn{1}{c|}{\multirow{2}{*}{\textbf{95.8\%}}} & \multirow{2}{*}{\textbf{46.5\%}} \\
 & \multicolumn{1}{c|}{} &  & \multicolumn{1}{c|}{} &  \\ \hline
\end{tabular}%
}
\caption{Ablation Study of Domain Knowledge Alignment. DivTOD w/o Align denotes the DivTOD without domain knowledge alignment.}
\label{Filtering mechanism}

\end{table}

\subsection{Quantity of Diverse Dialogues} 
\begin{figure}[t]
    \centering
    \begin{adjustbox}{minipage=\linewidth,scale=1.0}
    \subfigure[MWOZ]{
        \includegraphics[width=0.47\textwidth]{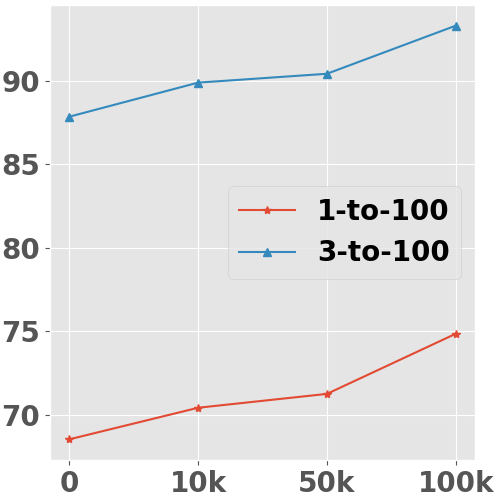}
    }
    \hspace{-0.4cm}
    \subfigure[DSTC2]{
        \includegraphics[width=0.47\textwidth]{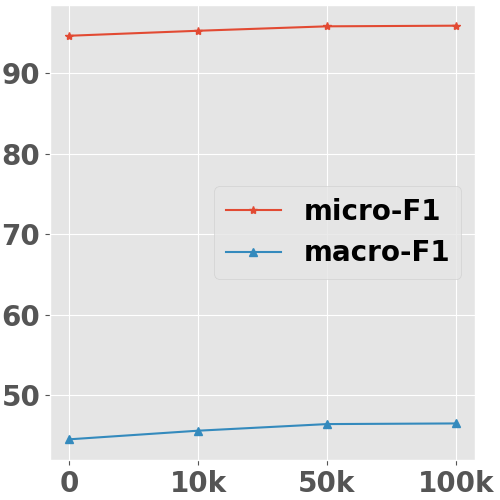}
    }

    \caption{The ablation experiment on the impact of the number of diverse dialogues generated by large language models on TOD.}
    \label{fig:ablation}
    \end{adjustbox}

\end{figure}

In our default experimental setting, we instructed LLMs to generate about 50k diverse dialogues for dialogue pre-training. Figure \ref{fig:ablation} shows the effect of varying the number of diverse dialogues during pre-training, on the two downstream tasks of dialogue act prediction on DSTC2 and response selection on MWOZ. We find the performance of DivTOD on both two tasks gradually improves as the number of diverse dialogues increases. This indicates that diverse dialogues generated by large language models can continuously improve the model's generalization.

\subsection{Few Shot Learning}

\label{sec:few_shot_setting}
Table \ref{few-shot} displays the results of dialogue act prediction on DSTC2 and response selection on MWOZ in few shot setting. \footnote{TOD-BERT uses the response contrastive loss as the pre-training objective on full MWOZ training data so we don't report its results on few-shot setting.} Our DivTOD achieves state-of-the-art results on all the metrics.  Although our DivTOD method and FutureTOD were both pre-trained using non-contrastive self-training frameworks, our DivTOD method significantly outperforms FutureTOD on both datasets in both 1\% and 10\% data settings. This demonstrates that DivTOD has a superior generalization capability across different scenarios.

We used the same pre-training corpus as the previous baseline, which includes MWOZ and CamRest676 (also part of MWOZ). However, concerns may arise about the reliability of few-shot performance due to this. To address this, we excluded MWOZ and CamRest676 from the pre-training corpus and evaluated the performance of FutureTOD and DivTOD on a 1\% data setting. On the MWOZ dataset, the performance of $\text{FutureTOD}^{\dagger}$ and $\text{DivTOD}^{\dagger}$ decreased. This suggests that including MWOZ in the pre-training phase does enhance few-shot performance. However, our $\text{DivTOD}^{\dagger}$ still achieves good performance and surpasses $\text{FutureTOD}^{\dagger}$. Surprisingly, on the DSTC2 dataset, $\text{DivTOD}^{\dagger}$ and $\text{FutureTOD}^{\dagger}$ also exhibit a slight decrease in performance. This highlights the high quality of MWOZ as a TOD dataset and provides some justification for its inclusion in the pre-training corpus in the previous baseline.

\begin{table}[t]
\centering
\resizebox{0.47\textwidth}{!}{
\begin{tabular}{|c|l|cc|cc|}
\hline
\multirow{2}{*}{\textbf{}} & \multicolumn{1}{c|}{\multirow{2}{*}{\textbf{Model}}} & \multicolumn{2}{c|}{\textbf{DSTC2}} & \multicolumn{2}{c|}{\textbf{MWOZ}} \\ \cline{3-6} 
 & \multicolumn{1}{c|}{} & \multicolumn{1}{c|}{\textbf{micro-F1}} & \textbf{macro-F1} & \multicolumn{1}{c|}{\textbf{1-to-100}} & \textbf{3-to-100} \\ \hline
\multirow{7}{*}{\textbf{1 \% Data}} & BERT & \multicolumn{1}{c|}{77.1\%} & 25.8\% & \multicolumn{1}{c|}{7.8\%} & 20.5\% \\
 & BERT-mlm & \multicolumn{1}{c|}{79.6\%} & 26.4\% & \multicolumn{1}{c|}{13.0\%} & 34.6\% \\
  & SimCSE & \multicolumn{1}{c|}{78.9\%} & 27.3\% & \multicolumn{1}{c|}{17.2\%} & 32.6\% \\
 & TOD-BERT & \multicolumn{1}{c|}{82.9\%} & 28.0\% & \multicolumn{1}{c|}{-} & - \\
 & DSE & \multicolumn{1}{c|}{72.4\%} & 21.4\% & \multicolumn{1}{c|}{7.9\%} & 21.2\% \\
 & $\text{FutureTOD}^{\dagger}$ & \multicolumn{1}{c|}{77.2\%} & 26.2\% & \multicolumn{1}{c|}{21.7\%} & 40.6\% \\
  & FutureTOD & \multicolumn{1}{c|}{83.7\%} & 31.0\% & \multicolumn{1}{c|}{35.8\%} & 53.5\% \\
 & $\text{DivTOD}^{\dagger}$ & \multicolumn{1}{c|}{79.0\%} & 26.9\% & \multicolumn{1}{c|}{24.6\%} & 45.2\% \\
 & DivTOD & \multicolumn{1}{c|}{\textbf{85.7\%}} & \textbf{36.5\%} & \multicolumn{1}{c|}{\textbf{36.9\%}} & \textbf{59.4\%} \\ \hline
\multirow{7}{*}{\textbf{10 \% Data}} & BERT & \multicolumn{1}{c|}{88.2\%} & 34.8\% & \multicolumn{1}{c|}{20.9\%} & 45.4\% \\
 & BERT-mlm & \multicolumn{1}{c|}{91.8\%} & 39.4\% & \multicolumn{1}{c|}{22.3\%} & 48.7\% \\
  & SimCSE & \multicolumn{1}{c|}{92.3\%} & 40.5\% & \multicolumn{1}{c|}{37.2\%} & 60.6\% \\
 & TOD-BERT & \multicolumn{1}{c|}{90.6\%} & 38.8\% & \multicolumn{1}{c|}{-} & - \\
 & DSE & \multicolumn{1}{c|}{91.1\%} & 39.0\% & \multicolumn{1}{c|}{24.8\%} & 49.4\% \\
 & FutureTOD & \multicolumn{1}{c|}{93.6\%} & 40.9\% & \multicolumn{1}{c|}{50.0\%} & 72.8\% \\
 & DivTOD & \multicolumn{1}{c|}{\textbf{95.1\%}} & \textbf{45.6\%} & \multicolumn{1}{c|}{\textbf{52.0\%}} & \textbf{76.5\%} \\ \hline
\end{tabular}%
}
\caption{Dialogue act prediction on DSTC2 and response selection on MWOZ for few-shot settings. $\text{DivTOD}^{\dagger}$ and $\text{FutureTOD}^{\dagger}$ are the models obtained by removing MWOZ and CamRest676 from the pre-training corpus. For DivTOD, we also excluded diversed dialogues generated from these two datasets.}
\label{few-shot}

\end{table}

\subsection{Zero Shot Learning} 
To validate the unsupervised embedding capability of our model, we performed zero-shot response selection on the MWOZ and DSTC2 datasets. The results are displayed in Table \ref{zero-shot}. BERT, FutureTOD, and DivTOD use an encoder architecture, while LLaMA and Vicuna use a decoder architecture. Therefore, for encoder models, we use the hidden state of the [CLS] layer as the embedding for inference retrieval, while for decoder models, we use the hidden state corresponding to the last input character as the embedding (consistent with the settings of DialoGPT). Our DivTOD outperforms BERT, FutureTOD, and LLaMA on all metrics, and is comparable to Vicuna. This indicates that the model has already gained strong context representation ability from the diverse dialogue data pre-training provided by Vicuna. However, the time cost and parameter size are much smaller than LLM like Vicuna, with a 14-fold and 70-fold reduction respectively


\begin{table*}[t]
\centering
\resizebox{0.75\textwidth}{!}{%
\begin{tabular}{|l|c|c|c|c|c|c|c|}
\hline
\multicolumn{1}{|c|}{\textbf{Model}} & \textbf{Dataset} & \textbf{1-to-100} & \textbf{3-to-100} & \textbf{5-to-100} & \textbf{10-to-100} & \textbf{Inference efficiency} & \textbf{Parameter size} \\ \hline
BERT & \multirow{5}{*}{\textbf{MWOZ}} & 1.8\% & 6.0\% & 9.9\% & 20.0\% & 89.9 & 110M \\
FutureTOD &  & 2.1\% & 6.3\% & 10.4\% & 20.7\% &89.9& 110M \\
LLaMA-7b &  & 2.2\% & 6.3\% & 10.4\% & 20.7\% &  5.3& 7B \\
Vicuna-7b &  & \textbf{2.6\%} & \textbf{7.3\%} & \textbf{11.8\%} & \textbf{22.7\%} &  5.3  & 7B \\
DivTOD (50k) &  & 2.5\% & 6.3\% & 11.0\% & 21.2\% & 89.9 & 110M \\ \hline
BERT & \multirow{5}{*}{\textbf{DSTC2}} & 1.0\% & 3.3\% & 6.3\% & 16.5\% & 89.9& 110M \\
FutureTOD &  & 1.7\% & 5.1\% & 9.0\% & 17.9\% & 89.9& 110M \\
LLaMA-7b &  & 2.1\% & 6.0\% & 10.1\% & 19.6\% &  5.3  & 7B \\
Vicuna-7b &  & 2.0\% & 6.2\% & 10.5\% & 20.4\% &  5.3  & 7B \\
DivTOD (50k) &  & \textbf{2.2\%} & \textbf{6.6\%} & \textbf{11.1\%} & \textbf{21.1\%} & 89.9 & 110M \\ \hline
\end{tabular}%
}

\caption{Response Selection on DSTC2 and MWOZ for zero-shot setting. We report 1-to-100, 3-to-100, 5-to-100, and 10-to-100 accuracy, which represents the ratio of the ground-truth response being ranked at the top-1,top-3, top-5 and top-10 given 100 candidates. Inference efficiency denotes the number of samples a model can infer per second when deployed on an Nvidia Tesla A100 GPU.}
\label{zero-shot}

\end{table*}

\subsection{Representation Diversity}
\label{sec:resp_diversity}

To understand whether our DivTOD can capture more diverse dialogue information and learn the intrinsic one-to-many diversity of TOD, we perform a qualitative analysis on the MWOZ test set. For each dialogue history, we select 2000 randomly sampled responses. We then compute the cosine distance between the representations of the dialogue history and response using a pre-trained response selection model in Table \ref{main_rs}.
We select the top 10 responses according to the cosine distance and compute Diversity and Coherence as the automatic metrics. Diversity denotes the number of unique types of dialogue acts in the top 10 responses. Coherence denotes average relevance scores between history and top-10 responses using a fine-tuned dual encoder in the response selection task.\footnote{We use TOD-BERT model in Table \ref{main_rs}, but we observe similar results using other response selection models.} We combine these two metrics to get the combined scores to measure the overall automatic response diversity and quality. The left part of Table \ref{ana_diver} shows the automatic results of different pre-trained models. Our model has advantages in all metrics, indicating that our model can capture rich dialogue policies without sacrificing response relevance. We also find TOD-BERT achieves comparable performance on coherence but performs worst on diversity, even worse than BERT. It proves that the noise introduced by the selection of positive and negative samples in contrastive learning may hurt the one-to-many diversity of dialogue representations.

Following \citet{zhang2020task}, we conduct a human evaluation to assess the appropriateness of individual responses and the diversity among selected responses. The appropriateness(App) is scored on a Likert scale of 1-3 for each response, while the diversity is scored on a Likert scale of 1-5 for all top 10 responses. We sample one hundred dialogue histories and corresponding top 10 responses retrieved by different pre-trained models. These samples are then scored by three judges given the dialogue history. The right part of Table \ref{ana_diver} shows the results of the human evaluation. We can find that the results of the human evaluation have the same trend as the automatic evaluation. Both the automatic evaluation and the human evaluation prove that our DivTOD model can learn the intrinsic one-to-many diversity of task-oriented dialogues.

\begin{table}[t]
\centering
\resizebox{0.45\textwidth}{!}{
\begin{tabular}{l|ccc|cc}
\hline
\multicolumn{1}{c|}{\multirow{2}{*}{Model}} & \multicolumn{3}{c|}{Automatic}                  & \multicolumn{2}{c}{Human}     \\ \cline{2-6} 
\multicolumn{1}{c|}{}                       & Diversity     & Conherence     & Combineed      & Diversity     & App           \\ \hline
BERT                                        & 5.50           & 0.668          & 12.18          & 1.97          & 1.23          \\
BERT-mlm                                    & 5.43          & 0.689          & 12.32          & 2.07          & 1.67          \\
SimCSE                                      & 5.48          & 0.675          & 12.23          & 1.93          & 1.47          \\
TOD-BERT                                    & 5.05          & 0.709          & 12.14          & 2.20           & 1.87          \\
DSE                                         & 6.17          & 0.680          & 12.97          & 2.53          & 1.33          \\
FutureTOD                                   & 6.33          & 0.706          & 13.39          & 2.67          & 1.91          \\
DivTOD                             & \textbf{7.92} & \textbf{0.730} & \textbf{15.22} & \textbf{2.88} & \textbf{1.94} \\ \hline
\end{tabular}
}

\caption{The automatic results and human evaluation results of response diversity on the MWOZ test set. The combined score is the overall automatic result which is calculated as follows: Combined score = Diversity + 10*Conherence.}
\label{ana_diver}

\end{table}





\section{Related Work}

\noindent\textbf{Dialogue Pre-trained Language Models} \citet{Zhang2020DIALOGPTL} use pre-trained GPT-2 model \cite{Radford2019LanguageMA} on Reddit data for open-domain dialogue response generation. PLATO \cite{Bao2019PLATOPD} pre-trains a dialog generation model with discrete latent variables using Twitter and Reddit data, which implicitly models dialog policy and solves the one-to-many mapping problem in open-domain dialog generation. However, since these models focus on chitchat dialogue, we do not compare them with our DivTOD. \citet{Wu2020TODBERTPN,Zhou2022LearningDR} use contrastive learning to learn TOD dialogue representations. \citet{Henderson2020ConveRTEA,Liu2021DialogueCSEDC} use similar ideas for dialogue retrieval and response selection. \citet{zeng2023futuretod} proposes a non-contrastive framework that distills future knowledge into the representation of the previous dialogue. Apart from these unsupervised methods, \citet{Zhou2022LearningDR,He2022SPACE2TS} use labeled dialogue data for supervised or semi-supervised pre-training. Since we focus on unsupervised TOD pre-training in this paper, we do not compare these models and leave it to future work.

\noindent\textbf{Enhancing small models with LLMs} Large Language Models (LLMs) \cite{han2021pre,bommasani2021opportunities}, such as ChatGPT and GPT-4 \cite{openai2023gpt4}, have demonstrated excellent generalization abilities in many language-related tasks. Recently, there have been many efforts to distill powerful LLMs for data augmentation, hoping to obtain equally powerful larger models through this approach without modifying the training objectives or model structures. For example, SelfInstruct \cite{wang2022self} and Alpaca \cite{Touvron2023LLaMAOA} generate 52k high-quality instruction-response pairs by distilling Text-Davinci-003, based on 175 seed tasks. In another line of work, LLMs are used to improve the ability of small models for specific tasks. \citet{ho2022large} and \citet{hsieh2023distilling} use LLMs to generate rationales that enhance the model's reasoning ability. \citet{liang2023let} uses LLMs as a math tutor to improve the model's math ability. In impossible distillation \cite{jung2023impossible}, LLMs help models generate high-quality and controllable summarizations and paraphrases. In contrast to previous work, we transfer rich background knowledge from LLMs to smaller models while filtering out domain knowledge that is irrelevant to the task-oriented dialogue system. 

\section{Conclusion}

We propose a new dialogue pre-training called DivTOD to diversify task-oriented dialogue representations by modeling the intrinsic one-to-many diversity of human conversations. DivTOD guides LLMs to transfer diverse background knowledge to smaller models while filtering domain knowledge that conflicts with task-oriented dialogues. Our experiments on various task-oriented dialogue tasks show that DivTOD outperforms FutureTOD, TOD-BERT, DSE, and other strong baselines. We plan to release all pre-trained models and code to facilitate future research. In the future, we hope to explore larger pre-trained models and more task-oriented dialogue corpora and extend similar ideas to generative dialogue models.

\section*{Limitations}
While DivTOD achieves significant improvements over existing baselines, there are still directions to explore for future work.
(1) We have designed a simple and effective method for LLMs to help dialogue pre-train models capture the intrinsic one-to-many diversity of human conversations. However, we have not considered solving this problem through the structure of the dialogue pretraining model. In the future, we will explore designing more efficient architectures for dialogue pretraining models and more efficient methods of knowledge transfer. (2) DivTOD only focuses on dialogue understanding tasks, such as dialogue act prediction and response selection. In the future, we will expand the idea of LLM collaborating with small models to generative dialogue pre-trained models. (3) We attempt various instructions to constrain the responses of $M_{T}$, including zero-shot prompts. However, these methods have not been very effective. For instance, the pass rate of the zero-shot method is relatively low in our post-filter. So we did not report these results. In the future, we plan to explore more advanced prompt techniques, such as the CoT method, to enhance our approach.

\section*{Ethics Statement}
We use Large Language Models (LLMs) to generate diverse responses. Despite our efforts to align domain knowledge, LLMs are inevitably prone to generating biased content. We anticipate that future research will focus on reducing the anti-social biases inherent in LLMs.

\section*{Acknowledgements}
We thank all anonymous reviewers for their helpful comments and suggestions. This work was supported by the National Natural Science Foundation of China (NSFC No.62076031 and No.62076036). This work is partially supported by State Key Laboratory of Massive Personalized Customization System and Technology (No. H\&C-MPC-2023-02-07(Q)). 


\bibliography{anthology,custom}
\bibliographystyle{acl_natbib}

\appendix

\section{Pre-training Data Statistics}

\label{sec:data_stastics}

We use the nine different task-oriented datasets collected by \cite{Wu2020TODBERTPN}: MetaLWOZ \cite{Lee2019MultiDomainTD}, Schema \cite{Rastogi2020TowardsSM}, Taskmaster \cite{Byrne2019Taskmaster1TA}, MWOZ \cite{Budzianowski2018MultiWOZA}, MSR-E2E \cite{Li2018MicrosoftDC}, SMD \cite{Eric2017KeyValueRN}, Frames \cite{Asri2017FramesAC}, WOZ \cite{Mrksic2017NeuralBT}, CamRest676 \cite{RojasBarahona2017ANE}. We show the full statistics in Table \ref{tb:train_dataset}.
\begin{table}[t]
\centering
\resizebox{0.98\linewidth}{!}{
\begin{tabular}{l|c|c|c|c}
\hline
\textbf{Name} & \textbf{ Dialogue} & \textbf{ Utterance} & \textbf{Avg. Turn} & \textbf{ Domain} \\ \hline
MetaLWOZ & 37,884 & 432,036 & 11.4 & 47 \\ \hline
Schema & 22,825 & 463,284 & 20.3 & 17 \\ \hline
Taskmaster & 13,215 & 303,066 & 22.9 & 6 \\ \hline
MWOZ & 10,420 & 71,410 & 6.9 & 7 \\ \hline
MSR-E2E & 10,087 & 74,686 & 7.4 & 3 \\ \hline
SMD & 3,031 & 15,928 & 5.3 & 3 \\ \hline
Frames & 1,369 & 19,986 & 14.6 & 3 \\ \hline
WOZ & 1,200 & 5,012 & 4.2 & 1 \\ \hline
CamRest676 & 676 & 2,744 & 4.1 & 1 \\ \hline
\end{tabular}
}
\caption{Data statistics for our pre-training task-oriented dialogue datasets.}
\label{tb:train_dataset}
\end{table}

\section{Baselines}
\label{sec:baseline}
DivTOD is evaluated on a variety of downstream tasks and compared to several well-established baselines. One such baseline is BERT \cite{devlin-etal-2019-bert}, which is the original BERT-base-uncased model that was pre-trained on a large text corpus. Another baseline is BERT-mlm, which is a version of BERT that underwent continual pre-training using MLM on our pre-training dialogue corpus. DialoGPT \cite{Zhang2020DIALOGPTL} is also included as a baseline, it is a decoder-only dialogue generation model that utilizes a language modeling target. SimCSE \cite{gao-etal-2021-simcse} constructs positive pairs using Dropout and undergoes further pre-training on the same TOD corpus. TOD-BERT \cite{Wu2020TODBERTPN} employs a contrastive response selection objective, treating a response utterance and its dialogue context as a positive pair. DSE \cite{Zhou2022LearningDR} takes consecutive utterances from the same dialogue as positive pairs.\footnote{In the interest of fairness, we use the unsupervised version of DSE, as done by \citet{zeng2023futuretod}.} FutureTOD\cite{zeng2023futuretod} uses a non-contrastive self-training framework with a self-distillation mechanism. It should be noted that some dialogue pre-training methods adopt an encoder-decoder architecture, but they usually use supervised settings, i.e. using labeled NLI datasets \cite{Williams2018ABC,Welleck2019DialogueNL} or dialogue act labels \cite{He2022SPACE2TS}. However, our focus is on unsupervised dialogue pretraining, and for fairness, we do not compare with them.




To validate the unsupervised embedding capability of our model, we also compared it with the 7B model LLaMA \cite{Touvron2023LLaMAOA} and Vicuna \cite{chiangvicuna} in a zero-shot response selection task. LLaMA is a powerful open-source large-scale model trained on a large corpus, while Vicuna is fine-tuned based on LLaMA using 70K high-quality conversation data.

\section{Implementation Details}
\label{implent_details}
\subsection{LLM generating Details}
\label{sec:llm_gen_detail}
We use Vicuna as LLM to generate diverse system responses and to align domain knowledge. For generation settings, the maximum generation length is 1024, the temperature is 0.7, and in order to ensure dialogue diversity, we choose to perform sampling. For verification settings, we obtain the logits corresponding to True and False in the first word of Vicuna’s decoding as the basis. In addition, if the model does not understand the task\footnote{For example, answering “I am a large language model” or “Okay, here is the written response you need”, etc.}, it will also be considered as not passing verification. If the response does not pass verification, Vicuna will generate a response again and verify it. The original dialogue will be retained if the generated response fails verification 5 times. Half responses in the dialogue will be rewritten. After diverse system response generation, all the dialogue will be merged into the original dataset as the new dataset to pre-train.

\subsection{Pre-training Details} 
\label{sec:pre_train_detail}
In DivTOD, we utilize a batch size of 48 and set the maximum input length to 512. The models are initialized using BERT-base-uncased and optimized using the Adam optimizer and a linear learning rate scheduler with an initial learning rate of 5e-5. A dropout ratio of 0.2 is employed and the mask ratio is set to 15\%. The predictor MLP head consists of two linear layers and a ReLU activation layer with an input dimension of 768 and a middle hidden dimension of 512. Upon completion of pre-training, all parameters of the Bert encoder are saved and the MLP head module is dropped for fine-tuning downstream tasks. Using an early-stopped strategy based on the perplexity scores of a held-out development, the pre-training process is conducted on eight NVIDIA Tesla A100 GPUs and takes five days. We use pre-trained models including BERT-MLM and TOD-BERT released by \cite{Wu2020TODBERTPN}, DSE model released by \cite{Zhou2022LearningDR}, and FutureTOD model released by \cite{zeng2023futuretod}. We re-implement SimCSE\cite{gao-etal-2021-simcse} using dropout to construct positive pairs and augment every single utterance obtained through dropout on our pre-training corpora. In terms of computational efficiency during pre-training, our DivTOD model is comparable to other baselines.

\subsection{Finetuning Details}
\label{sec:fine_tune_detail}
We directly use the results reported by TOD-BERT \cite{Wu2020TODBERTPN} for BERT-mlm and TOD-BERT. We adopt the same hyperparameters for all downstream tasks except the batch size and learning rate. We finetune all downstream tasks with the original dataset for 50 epochs with an early-stopped strategy evaluated on the validation set every 100 steps with patience set to 10. We respectively set the batch size to 8, 25, 16, and 100 for intent recognition, dialogue state tracking, dialogue act prediction, and response selection and keep the learning rate to 5e-5 for all the tasks. 

\section{Prompt Examples}

\begin{figure}[t]
 \centering
\resizebox{0.45\textwidth}{!}{
 \includegraphics[scale=0.5]{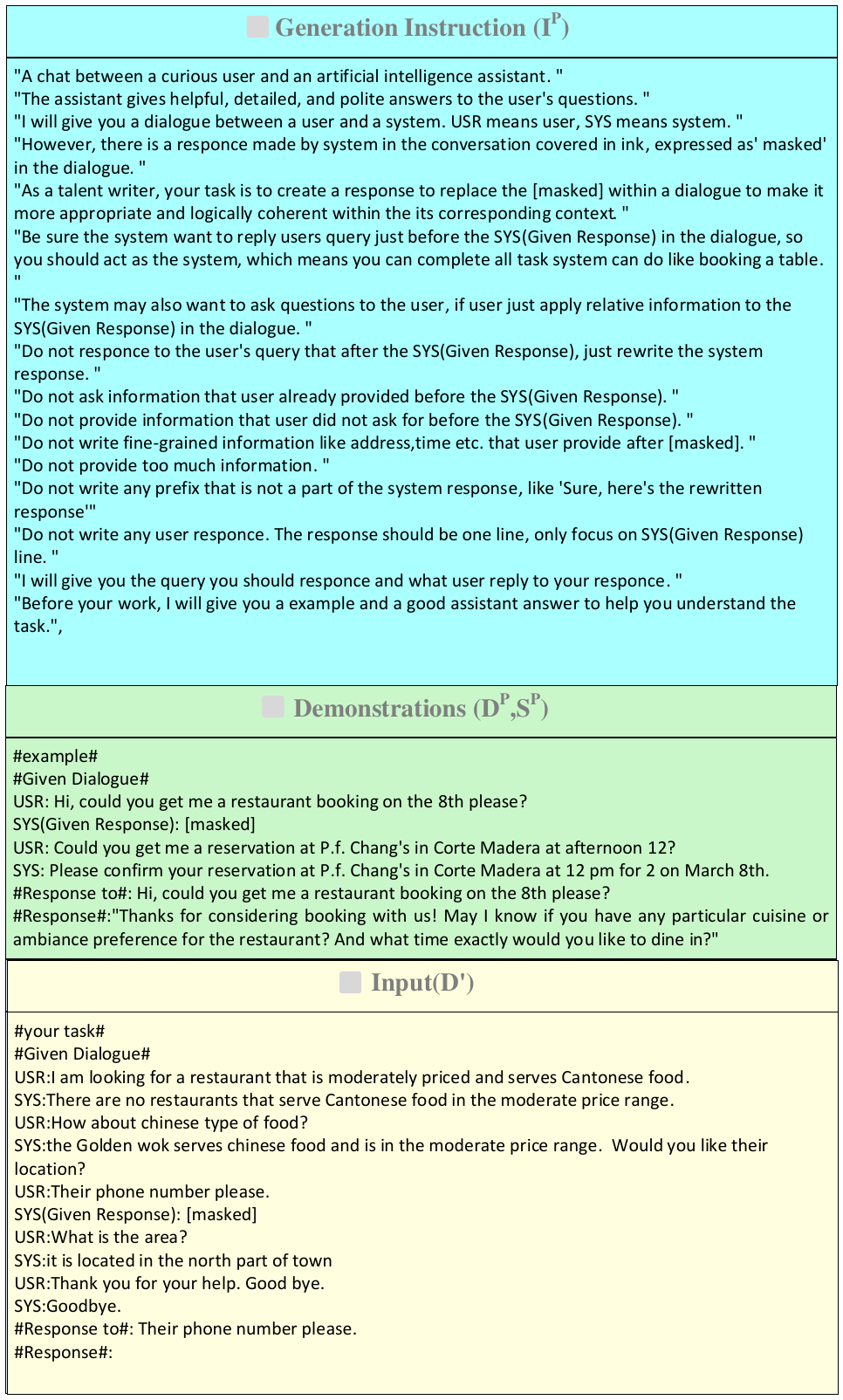}
 }
\vspace{-0.3cm}
 \caption{The complete prompt example for generating diversified responses.}
 \label{fig:prompt_gen}
 \vspace{-0.3cm}

\end{figure}

\begin{figure}[t]
 \centering
\resizebox{0.5\textwidth}{!}{
 \includegraphics[scale=0.5]{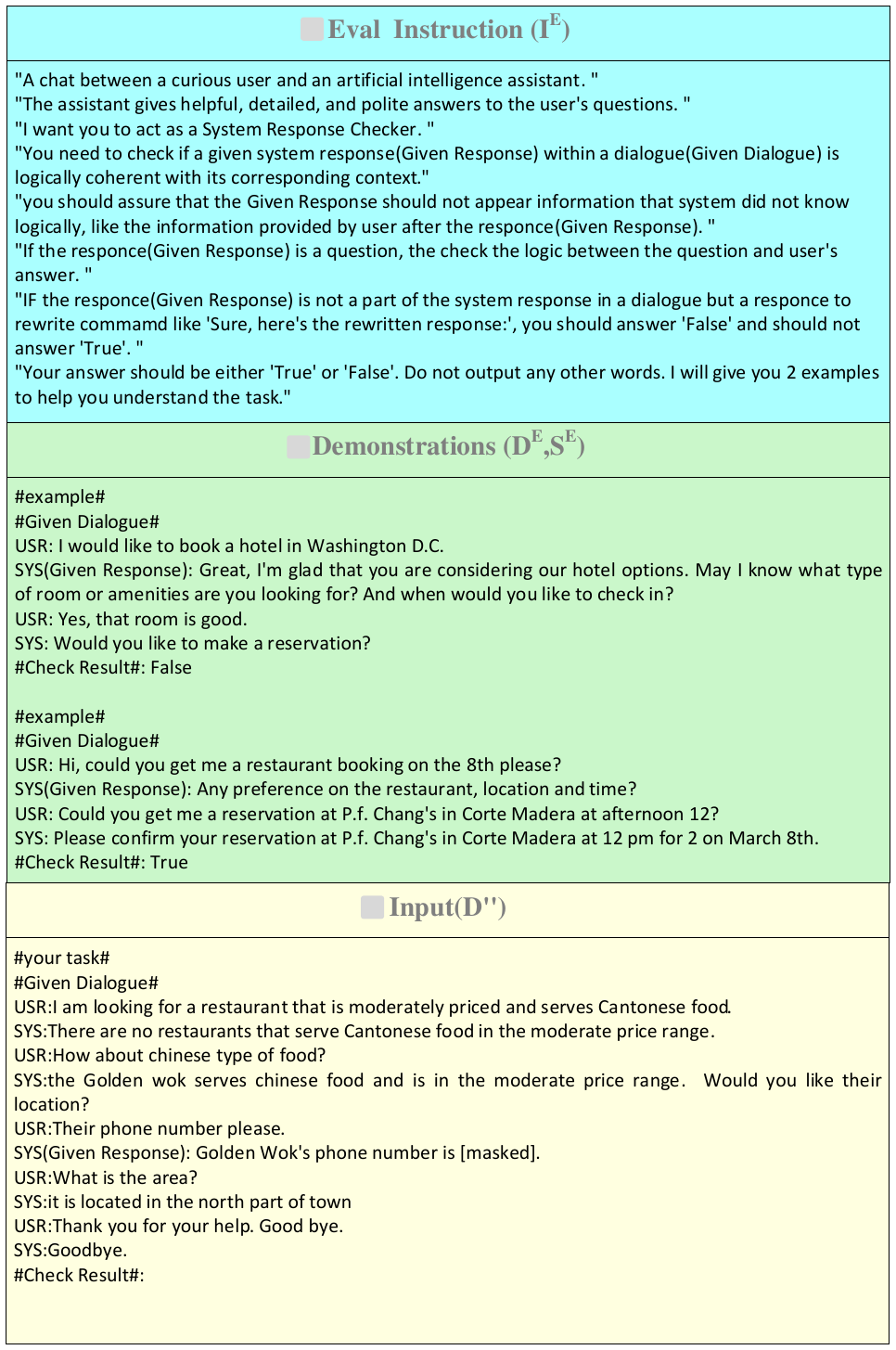}
 }
\vspace{-0.3cm}
 \caption{The complete prompt example for aligning domain knowledge.}
 \label{fig:prompt_eval}
 \vspace{-0.3cm}

\end{figure}

We provided prompts for generating diversified responses and aligning domain knowledge in Figure \ref{fig:prompt_gen} and Figure \ref{fig:prompt_eval}, respectively.

\appendix

\begin{algorithm}
	\renewcommand{\algorithmicrequire}{\textbf{Input:}}
	\renewcommand{\algorithmicensure}{\textbf{Output:}}
	\caption{Generating-Aligning Steps}
	\label{alg1}
	\begin{algorithmic}[1]
		\STATE \textbf{Initialization:} Generation Prompt with Example $P$, Eval Prompt with Example $E$, Try Turns $T$, Model $M_T$ 
            \STATE \textbf{Input:} a Dialogue $D$, Dialogue Turns Number $n$
                \FOR{N in [1,$\left \lfloor n/2 \right \rfloor$]}
                    \STATE $try\_number$ = 0, $filtering\_result$=False
                    \WHILE{$try\_number$ <  $T$}
            		  \STATE $D'$ = Replace $S_i$ in $D$ into [masked]
                        \STATE $S_i'$ = $M_T$($P$, $D'$)
                        \STATE $D''$ = Replace $S_i$  in $D$ into $S_i'$
                        \STATE $filtering\_result$ = $M_T$($E$, $D''$)
                        \STATE $try\_number$ += 1
                        \IF{$filtering\_result$ is True}
                            \STATE $D$ = $D''$
                            \STATE \textbf{break}
                        \ENDIF
                    \ENDWHILE
            \ENDFOR
		\ENSURE  $D$
	\end{algorithmic}  
 \label{algo}
\end{algorithm}

\end{document}